\documentclass[11pt,a4paper]{article}
\usepackage[hyperref]{acl2020}
\usepackage{times}
\usepackage{amsmath}
\usepackage{amsmath}
\usepackage{array}
\usepackage{subfigure}
\usepackage{graphicx}
\usepackage{latexsym}

\usepackage{microtype}

\aclfinalcopy % Uncomment this line for the final submission
\title{A Question Type Driven and Copy Loss Enhanced Framework \\
for Answer-Agnostic Neural Question Generation}

\author{Xiuyu Wu\textsuperscript{1}, Nan Jiang\textsuperscript{2}, Yunfang Wu\textsuperscript{2}\thanks{\quad Corresponding author.} \\
\\ \textsuperscript{1}School of Foreign Language, Peking University \\
\textsuperscript{2}MOE Key Lab of Computational Linguistics, School of EECS, Peking University \\
texttt{xiuyu\_wu@pku.edu.cn,jnhsyxxy@126.com,wuyf@pku.edu.cn}
\\}
\date{}

\begin{document}
\maketitle

\begin{abstract}
The answer-agnostic question generation is a significant and challenging task, which aims to automatically generate questions for a given sentence but without an answer. In this paper, we propose two new strategies to deal with this task: question type prediction and copy loss mechanism. The question type module is to predict the types of questions that should be asked, which allows our model to generate multiple types of questions for the same source sentence. The new copy loss enhances the original copy mechanism to make sure that every important word in the source sentence has been copied when generating questions. Our integrated model outperforms the state-of-the-art approach in answer-agnostic question generation, achieving a BLEU-4 score of 13.9 on SQuAD. Human evaluation further validates the high quality of our generated questions. We will make our code public available for further research.  
\end{abstract}

\section{Introduction}
\noindent Question Generation (QG) has been investigated for many years because of its huge potential benefits to various fields, especially for education \cite{Mitkov2006A,Rus2009THE,Heilman2010Good}.
% since questions are tools to identify whether students have mastered the knowledge. 
%To reduce the burden of educational experts, automatic question generation has been proposed as a machine learning task for decades \cite{Mitkov2006A,Rus2009THE}. 
QG can also act as an essential component of other comprehensive task, such as dialogue systems \cite{Piwek2007T2D,Shum2018From}. Besides, it can supplement question answering task by automatically constructing a large question set \cite{Duan2017Question,Tang2017Question}.

Traditional methods for automatic question generation are mostly rule-based, which need a set of complex empirical rules \cite{Beulen1998Automatic,Brown2005Automatic,Heilman2010Good,Mazidi2014Linguistic}. Recently, with the flourish of deep learning, especially the sequence-to-sequence (seq2seq) frame \cite{Sutskever2014Sequence} with attention mechanism \cite{Bahdanau2014Neural}, many neural models have been proposed to solve the QG task and achieve rapid progress \cite{Du2017Learning,Zhou2017Neural,Tang2017Question,Tong2017A,Zhao2018Paragraph,Dong2019Unified,Nema2019Let,Zhou2019Multi}. 

\begin{table}[t]
\small
\begin{tabular}{p{7.2cm}}
\hline
\textbf{Source sentence}: the notion of style in the arts was not developed until the 16th century , with the writing of vasari : by the 18th century, his lives of the most excellent painters , sculptors, and architects had been translated into italian , french , spanish and english . \\
\hline
\hline
\textbf{Groud-truth}: \\
Q1: \textbf{when} were the \emph{styles} of \emph{arts} created ?  \\
Q2: \textbf{who} wrote \emph{lives} of the most \emph{excellent painters} , \emph{sculptors} , and \emph{architects} ?  \\
Q3: by the \emph{18th century} \textbf{which} languages was \emph{vasaris} book \emph{translated} in ?  \\
Q4: in \textbf{what} \emph{century} did `` \emph{style} '' as an artistic concept arise ? \\
\hline
\end{tabular}
\caption{Different types of questions with respect to the same source sentence, where the italics words are keywords occurring both in the source sentence and reference questions.}
\label{multi-types example}
\end{table}

However, most of the previous works are devoted to deal with answer-aware question generation. That is, given a text and also an answer span, the system is required to generate questions. But in a real application for educational purpose, people or machines are often required to generate questions for natural sentences without explicitly annotated answer. Comparing with the answer-aware QG, the answer-agnostic QG (AG-QG) task is more challenging and attractive. Unfortunately, AG-QG has been much less studied. Du's work \cite{Du2017Learning} is the first one to tackle this problem, and \cite{Scialom2019Self} achieve the state-of-the-art by employing an extended transformer network. 
% with a sequence-to-sequence attention model, and now the state-of-the-art result in answer-agnostic QG task is achieved by the extended transformer network \cite{Scialom2019Self}. 
% There is much room for improvement in this task. 

For the AG-QG task, where the input is only a sentence but without any answer, multiple questions might be asked from various perspectives. According to our statistics on SQuAD \cite{Rajpurkar2016SQuAD}, 
%which is the most commonly used dataset for QG, we find that 
nearly 34$\%$ of the source sentences are offered multiple gold reference questions, and nearly 20$\%$ of the source sentences are offered different types of questions. Table \ref{multi-types example} gives an example, where one source sentence corresponds with four different types of questions. However, most existing approaches can only generate one question for one input sentence. 
 
%To address the multi-types QG and to ask reasonable questions 
%For a natural sentence that generally expresses the meaning of "who does what to whom when where how", the most important thing to ask reasonable questions is to determine what types of questions should be asked, i.e., which part of information should be focused on when asking questions. 
To enable the model to ask different types of questions given the same input sentence,
we propose a question type driven framework for AG-QG task. Specially, our model firstly predicts the probability of different question types distribution on the input sentence, 
%via a multi-task learning with the main question generation task. The model 
which allows us to choose the best $K$ question types with higher possibility. Then these different question types will be embedded into different vectors, which will 
%teach the model to learn the specific inner pattern of different types and 
guide the decoder to pay attention to informative parts with respect to different questions.   

Meanwhile, according to the statistics on SQuAD, on average there are 3.09 non-stop words copying from the source sentence for each reference question. Those non-stop words appearing in both questions and sentences are regarded as keywords, since they act as the connection of these two parts. To increase the probability of copying keywords from source sentences, we design a new copy loss to enhance the traditional copy mechanism. 
%Since the traditional loss only cares about whether the $i_th$ generated word is exactly the same as in the golden question, but never pays attention to whether these keywords have been copied into the generated question. 
In our model, by minimizing the new copy loss, the model will be forced to copy these keywords at least once during decoding.  
% We design this copy loss to let the highest copy probability of key words to be as near to 1 as possible. That is, 

We conduct experiments on SQuAD. Both the question type module and the new copy loss improve performance over the baseline model, and our full model combining two modules obtains a new state-of-the-art performance with a BLEU-4 of 13.9. Moreover, our model can ask different types of questions for a given sentence. 
% just when the human-assigned parameter $K$ is set to more than one.  
% To improve the performance of our model, we enhance the traditional copy mechanism in order to make sure all these important key words are copied. Specially, we design a copy loss to increase the probability of copying key words from source sentences. 

% Another benefit brought by question type embedding module is that, by assigning different question types, the model can generate different questions even though the input source sentence is the same. That is to say, the question type embedding module endows our model the ability to generate multiple questions for the same source sentence, which is more natural and conform to reality, since in real setting, different questions about one passage might be asked from various perspective. The point of our question type predict part is that although the model have the ability to generate all types of questions for one input, we still want it to find the most possible, the most proper and the best question to generate.

%In sum, this paper makes the following contributions:
We conclude the contributions as follows: 

\begin{itemize}
\item We propose a question type driven framework for AG-QG, which enables the model to generate diverse questions with high quality. 

% predict module to learn given the source sentence, which types of questions should be generated. And we use a question type embedding module to improve the performance of generating certain types of questions.
% We also a question type predict module to predict the most proper type of the question that would be generated.
\item We design a new copy loss function to enhance the standard copy mechanism, which increases the probability of copying keywords from source sentences.
\item Our model achieves a new state-of-the-art performance for the challenging AG-QG. The human evaluation further validates a high quality of our generated questions in fluency, relevance and answer-ability. 
% and it can ask multiple types of questions with respect to a same input sentence.    

\end{itemize}

\section{Related Work}

\paragraph{Answer-Aware Question Generation.} 
Most previous works on question generation focus on answer-aware QG task. Yuan \cite{Yuan2017Machine} proposes three parts of loss to enhance the performance of sequence-to-sequence attention model. Zhou \cite{Zhou2017Neural} leverage lexical features (part-of-speech and named entity tags) to help the model get better encoder representation. Zhao \cite{Zhao2018Paragraph} uses paragraph information to do answer-aware QG task. And to use answer information more efficiently, Song \cite{Song2018Leveraging} uses multi-perspective matching and Sun \cite{Sun2018Answer} proposes position-aware model to pay more attention to the surrounding context of answer span. \cite{Nema2019Let} uses a answer encoder to encode answer and fusion it with paragraph representation. \cite{Chen2019Natural} applies reinforcement learning to increase the performance. 

\paragraph{Answer-Agnostic Question Generation.} 
AG-QG is more challenging than answer-aware QG. Du's work \cite{Du2017Learning} is the first one to tackle this problem, and they achieve better performance than rule-based approaches by employing a sequence-to-sequence attention model. \cite{Du2017Identifying} aim to automatically find question-worthy sentences from a paragraph and then generate questions. \cite{Subramanian2017Neural} treat QG as a two-stage task: answer phrase extraction and answer-aware question generation. \cite{Wang2019A} propose a multi-agent communication framework, using a local extraction agent to extract question-worthy phrases, and then taking extracted phrases as assistance to generate questions.  \cite{Scialom2019Self} employ the transformer network \cite{Vaswani2017Attention} and extend it with the placeholder strategy, copy mechanism and contextualized embedding.

\paragraph{Question Word Prediction.}
Question word is one of the most important components of a question. \cite{Fan2018A} study multi-types visual question generation, by feeding the encoded representation to a multi-layer perception to calculate the question words distribution. \cite{Sun2018Answer} propose an answer-focused and position-aware model to generate the first question word. \cite{Kim2018Improving} propose an answer-separated sequence-to-sequence model to identify the proper question word. They replace the answer span in the source sentence with a special token to make better use of the context information.

\paragraph{Multi-Types Question Generation.} 
The multi-types QG has been much less researched. In Ma's work \cite{Ma2018Aspect}, in order to generate different types of questions, they use question type embedding at the first step of decoding. However, because of the difficulty in automatically predicting question types, their model fails to outperform the previous works. The question type driven framework has also been tried for visual question generation \cite{Fan2018A}, where they concatenate the question type embedding with the encoded representation of input. 
% Although some progress has been made, there is still much room for improvement for multi-types QG.

\section{Proposed Framework}

\subsection{Framework Overview} For each sample in our dataset, we have a source sentence $S=(x_1,x_2,\ldots,x_l)$, which is a word sequence and \textit{l} denotes the number of words. Let $Q=(y_1,y_2,\ldots,y_m)$ to represent the question, which is another word sequence and \textit{m} is the length. The answer-agnostic QG task can be defined as finding the best $\overline{\emph{Q}}$ that:

\begin{displaymath} 
    \begin{aligned}
	\overline{Q} & = 
	\mathop{\arg\max_{Q}} \log{P}(Q \mid S)  \\
	& = \mathop{\arg\max_{Q}} \sum_{i=1}^{m}\log{P}
	(y_{i} \mid S,y_{<i})
	\end{aligned}
\end{displaymath} 

\begin{figure*}[pt]
	\centering
	\includegraphics[width=\linewidth]{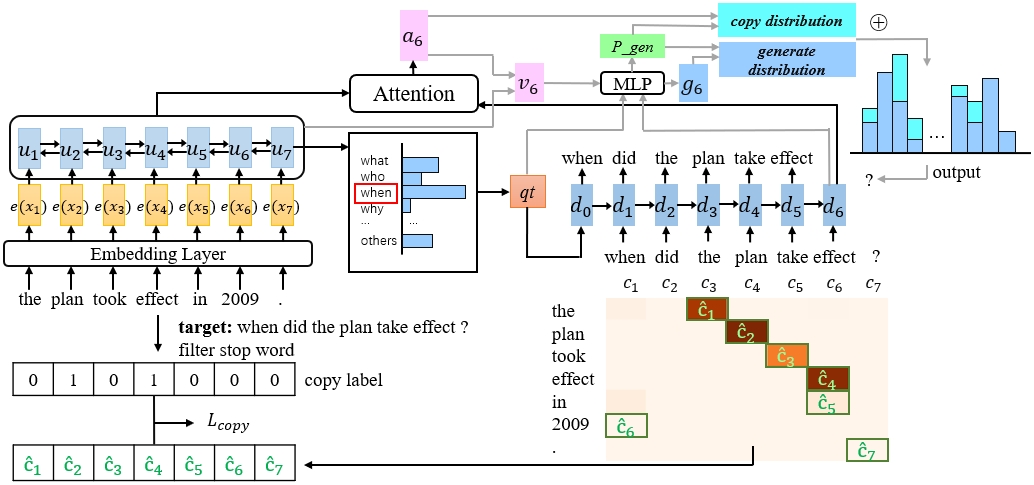} 
	\caption{Our copy loss enhanced and question types driven framework}
	\label{model framework}
\end{figure*}

Figure \ref{model framework} shows the framework of our full question type driven QG model. With the development of pointer network \cite{Vinyals2015Pointer}, the copy mechanism \cite{Gu2016Incorporating} has been increasingly more applied to natural language generation task. Thus, our model is based on the general sequence-to-sequence attention model with copy mechanism, which we regard as our baseline. In the next sections, we firstly describe the baseline model and then separately show our question type module and enhanced copy mechanism. 

\subsection{Baseline Model}
Our baseline method is a sequence-to-sequence attention model with copy mechanism. 
%The encoder  consists of an embedding layer and a bi-directional LSTM layer. 
Let $x_t$ represent the $t-th$ word in the source sentence and $e(x_t)$ is its corresponding embedded vector. A bi-directional LSTM layer \cite{Graves1997Long} is used to encode the embedded vector sequence: 
% $\mathbf{E}(S)=(e(x_1),e(x_2),\ldots,e(x_l))$:

\begin{equation}
    \begin{aligned}
    \overrightarrow{u_t} & = \overrightarrow{LSTM}(\overrightarrow{u_{t-1}}, e(x_t)) \\
    \overleftarrow{u_t} & = \overleftarrow{LSTM}(\overleftarrow{u_{t+1}},e(x_t)) \\
    u_t & = [\overrightarrow{u_t};\overleftarrow{u_t}]
    \end{aligned}
\end{equation}

The hidden state at time step $t$ is the concatenation of the forward $\overrightarrow{u_t}$ and backward $\overleftarrow{u_t}$,  which can be represented as $\mathbf{U} = {[\overrightarrow{u_t};\overleftarrow{u_t}]}_{t=1}^{l}$, where $l$ is the number of words in the source sequence.

The decoder is another LSTM network, which generates a new hidden state $h_t$, conditioned on the previous state $h_{t-1}$ and previously embedded generated word $e(y_{t-1})$. At the first decoding step, it takes the last encoding hidden state $u_l$ and a special token $[SOS]$, which stands for the start of sequence, as input:

\begin{equation}
    \begin{aligned}
    h_t & = LSTM(h_{t-1}, e(y_{t-1})) \\
    h_0 & = LSTM(u_l, e([SOS]))
    \end{aligned}
\end{equation}

Given the encoded state $\mathbf{U}$ and decoder state $h_t$, the baseline model calculates a generating distribution of words at each time step $t$, which is calculated as follow:

\begin{equation}
    \begin{aligned}
    g_{t} & = softmax(\mathbf{W^o}tanh(\mathbf{W^t}[h_t;v_t]))  \\
    v_{t} & = \sum_{i=1}^{l}a_{it}u_i  \\
    a_{it} & = \frac{\exp(h_t^\top \mathbf{W^b}u_i)}{\sum_{j}\exp(h_t^\top \mathbf{W^b} u_j)}
    \end{aligned}
\end{equation}
where $v_t$ is the weighted sum of the encoded representation $\mathbf{U}$, and the attention weight $a_t$ is calculated by a bi-linear scoring function and a softmax normalization. $\mathbf{W^t}$ and $\mathbf{W^o}$ are both trainable parameters, which can be regarded as applying a multi-layer perceptron to the concatenation of hidden state $h_t$ and global attention representation $v_t$. $\mathbf{W^b}$ is also trainable parameter that is applied to calculate attention weights.

Our baseline model also utilizes the copy mechanism. We use the attention score $a_t$ obtained from the decoder attention as the copy distribution. The probability of generating a word, $p\_gen$, is calculated as $p\_gen = sigmoid(\mathbf{W^g}[h_t;v_t])$. The final distribution is the weighted sum of generating distribution and copy distribution:

\begin{equation}
    \begin{aligned}
    \hat{g}_t & = p\_gen \cdot g_t \\
    c_t & = (1-p\_gen) \cdot a_t
    \end{aligned}
\end{equation}
where $\hat{g}_t$, $c_t$ are the weighted generate distribution and copy distribution, respectively. We use $f_t$ to represent the final distribution and $f_{ti}$ is the probability of decoding the $i-th$ word in the vocabulary. Let $w_i$ represent the $i-th$ word in the vocabulary. The final distribution is calculated as follow:

\begin{displaymath}
    f_{ti} = 
        \left\{ \begin{array}{ll} \hat{g}_{ti} + \sum\limits_{k,where \ x_k=w_i}c_{tk} & w_i \in \mathbf{X}  \\ 
        \hat{g}_{ti} & otherwise
        \end{array} \right. 
\end{displaymath}

Given the training corpus $D$, in which each sample contains a source sentence $S$ and a target question $Q$, the training objective is to minimize the negative log-likelihood of the target questions $L$:

\begin{equation}
    L = - \sum_{<S,Q>\in D}\log{P}(Q|S;\theta)
\end{equation}
where $\theta$ represents all the parameters of our model. 

\subsection{Question Type Prediction}
We propose the question type module for two goals. One is to enable our model to generate multiple types of questions for one source sentence, and the other is to improve the generating performance. Our question type module firstly predicts the most proper type and then uses the embedding of it to help the decoding process.

As for question types, we count the distribution of question types in SQuAD and finally category all the questions into 7 types: what, who, how, where, when, yes/no and others.

The question type prediction is a multi-layer perceptron, which takes the last hidden state of encoder %(the average of all the hidden states is also tried but gets same performance) 
as input to predict the probability distribution of question types, denoted by $\mathbf{T}$:

\begin{equation}
    \mathbf{T} = softmax(MLP(u_l))
\end{equation}

Please note that our model can generate multiple questions for one source sentence. When the number of questions need to be generated is set to $K$, our model will select $K$ question types with the highest probability as output. Consequently, our decoder will decode $K$ times. 

\begin{equation}
    ty_1, ty_2, \ldots, ty_K = TopK(\mathbf{T})
\end{equation}

At every decoding time, for one of the best $K$ question types, $ty$, we embed it into a question type vector $qt$:

\begin{equation}
    \begin{aligned}
    qt & = Embedding(ty) \\
    ty & \in {ty_1,ty_2,\ldots,ty_K}
    \end{aligned}
\end{equation}

The embedded question type vector will be used in decoding, and in this way, the question type vector would guide the model to generate questions that follow the pattern of a specific question type. Specially, we use the question type vector $qt$ instead of the embedding of $[SOS]$ token as the input of the decoder at the first decoding step:

\begin{equation}
    h_0 = LSTM(u_l, qt)
\end{equation}

Besides, when calculating the generating distribution $p\_gen$, we concatenate $qt$ with the global attention representation and the decoder hidden state:

\begin{equation}
    \begin{aligned}
    & g_{t} = softmax(\mathbf{W^o}tanh(\mathbf{W^t}[h_t;v_t;qt])) \\
    & p\_gen = sigmoid(\mathbf{W^g}[h_t;v_t;qt])
    \end{aligned}
\end{equation}

Finally, we train the question type prediction and question generating simultaneously in multi-task learning framework. For each sample $<S,Q>$, we calculate the ground-truth question type distribution $TY$ and we add an additional factor to the loss function, which is the negative log-likelihood of the target questions' types:

% For each sample $<S,Q>$, we use question word matching method to get the question type of $Q$, namely $ty$. We train the question generating part and question type prediction part simultaneously, by  

\begin{equation}
    \begin{aligned}
    L = - \sum_{<S,Q,TY>\in D}[ & \log{P}(Q|S;\theta)+ \\
    & \lambda_1\log{P}(TY|S;\theta)]
    \end{aligned}
\end{equation}
where $\lambda_1$ is a hyper-parameter to balance two parts. 

\subsection{Enhanced Copy Mechanism}
According to our observation on SQuAD, when creating questions people often (may have to) copy some keywords from the source sentence. So we consider non stop words appearing in both questions and source sentences as keywords, and we assume that these keywords should also occur in the generated sequence. In order to push the model to copy keywords from source sentences, we propose a copy loss to enhance the traditional copy mechanism. For a word $x_l$ in the source sentence, we first define a function $cl(copy label)$ as follow:

%\begin{displaymath}
\begin{equation}
    \begin{aligned}
    cl(x_i) = 
        \left\{ \begin{array}{ll} 1 & x_i \in Q\ and\ x_i \notin stop word \\ 
        0 & otherwise
        \end{array} \right.  \\
    \end{aligned}
\end{equation}
% \end{displaymath}
We believe that at one decoding step, the copy probability of keywords ($cl(x_i)=1$) should be close to 1. So we define the copy loss as:

\begin{equation}
    \begin{aligned}
    & \hat{c}_i = max(c_{1i},c_{2i},\ldots,c_{mi})& \\
    & L_{copy} = \sum_{i=1}^{l}cl(x_i)(\hat{c}_i - 1)^2 
    \end{aligned}
\end{equation}
where $c_{ti}$ is the copy probability of the $i-th$ word in the source sentence at the $t-th$ decoding step, computed by Equation 4. Thus, $\hat{c_i}$ is the highest copy probability of the $i-th$ word in the source sentence among all the $m$ decoding step. $l$ denotes the number of words in the source sentence.

Finally, we add the copy loss into the total loss:

\begin{equation}
    \begin{aligned}
    L = & - \sum_{<S,Q,TY>\in D}[\log{P}(Q|S;\theta) \\
    & +\lambda_1\log{P}(TY|S;\theta) \\
    & +\frac{\lambda_2}{l}\sum_{i=1}^{l}cl(x_i)(\hat{c}_i - 1)^2]
    \end{aligned}
\end{equation}
where $\lambda_2$ is another hyper-parameter to control the impact of penalty loss. 

\section{Experimental Settings}
\subsection{Dataset and Pre-processing}
We conduct experiments on the SQuAD dataset \cite{Rajpurkar2016SQuAD}, which contains more than 70k training samples, 10k development samples and 11k test samples. In either training, development or test dataset, multiple samples might share the same source sentence but with different target questions. But a same source sentence will not appear in different datasets, which ensures the confidentiality of test data.

% \subsection{Data Preparation}
We adopt subword representations \cite{Sennrich2015Neural} rather than raw words, which can not only reduce the size of vocabulary, increase the training speed, address the problem of out of vocabulary words, but also improve the model performance. By using byte-pair encoding, our vocabulary size is reduced to less than 6k.  Due to the vanishing gradient problem in recurrent neural networks \cite{pascanu2013difficulty,hochreiter1998vanishing}, we choose 256 for the maximum length of inputs and 50 for the maximum length of target questions. 

\subsection{Implementation Details}
We adopt a 2-layers bi-directional LSTM for encoding and a 1-layer LSTM for decoding. The number of hidden units is 600, and the dimension of both word embedding and question type embedding is 300. We do not use pre-trained word embedding since we use subword representations rather than word-level representations. The drop rate \cite{Krizhevsky2012ImageNet} between each layer is 0.3. We firstly use Adam \cite{Kingma2014Adam} with learning rate of 0.001 for fast training, and after training 5 epochs, the stochastic gradient descent(SGD) with learning rate of 0.01 is used for fine-tuning. We train our model for 15 epochs with mini-batch size of 64. During training, hyper-parameter K is set to 1 and when decoding, we do beam search with a beam size of 4. For hyper-parameters $\lambda_1$ and $\lambda_2$, we try different settings and choose the best one by observing the descending trend of total loss and ascending trend of BLEU-4 score on the valid set. Finally, both $\lambda_1$ and $\lambda_2$ are set to 0.1.

\subsection{Evaluation Metrics}
We adopt BLEU \cite{Papineni2002BLEU}, METEOR \cite{Denkowski2014Meteor} and ROUGE-L \cite{Flick2004ROUGE} for evaluation, and use the evaluation package released by Chen \cite{Chen2015Microsoft}. BLEU measures the precision of n-grams on a set of references, with a penalty for over short generation. METEOR calculates the similarity between generations and references by considering synonyms, stemming and paraphrases. ROUGE measures the recall of n-grams on the set of references. 

\begin{table*}[t!]
\small
\centering
\begin{tabular}{lcccccc}
\hline
\bf model & \bf BLEU-1 & \bf BLEU-2 & \bf BLEU-3 & \bf BLEU-4 & \bf METEOR & \bf ROUGE-L \\
\hline
\hline
Seq2SeqAtt & 42.90 & 25.97 & 17.68 & 12.49 & 16.85 & 39.59 \\ 
Seq2SeqAtt (Du) & 43.09 & 25.96 & 17.50 & 12.28 & 16.62 & 39.75  \\
Transformer (Scialom) & 43.33 & 26.27 & 18.32 & 13.23 & - & 40.22  \\
\hline
Our model & \textbf{45.08} & \textbf{27.98} & \textbf{19.38} & \textbf{13.90} & \textbf{18.12} & \textbf{40.77} \\
\hline
\end{tabular}
\caption{Experimental results comparing with previous works.}
\label{evaluation of different models}
\end{table*}

\begin{table*}[t!]
\small
\centering
\begin{tabular}{lcccccc}
\hline
\bf model & \bf BLEU-1 & \bf BLEU-2 & \bf BLEU-3 & \bf BLEU-4 & \bf METEOR & \bf ROUGE-L \\
\hline
\hline
Baseline & 43.39 & 26.75 & 18.48 & 13.17 & 17.40 & 40.57  \\
Baseline+Type & 44.59 & 27.25 & 18.76 & 13.41 & 17.54 & 40.41 \\
Baseline+CopyLoss & 44.45 & 27.62 & 19.08 & 13.65 & 17.90 & 40.16 \\
Baseline+CopyLoss+Type & \textbf{45.08} & \textbf{27.98} & \textbf{19.38} & \textbf{13.90} & \textbf{18.12} & \textbf{40.77} \\
\hline
Upper Bound & 48.42 & 30.28 & 21.12 & 15.27 & 19.43 & 43.82 \\
\hline
\end{tabular}
\caption{Ablation study of our model.}
\label{ablation study}
\end{table*}

\section{Results and Analysis}
In this section, we report the automatic evaluation results of our proposed model and do ablation study to prove the effectiveness of different parts of the model. Then we conduct human evaluation and case study to test the quality of generated questions. Furthermore, we give a detailed analysis on multiple questions generation.    

\subsection{Main Results}
We compare our model with the following previous works:
\begin{itemize}
\item \textbf{Seq2Seq Attention}: It is a traditional sequence-to-sequence attention model.
\item \textbf{Seq2SeqAtt (Du)}: This is the first work in AG-QG task, which is a sequence-to-sequence attention model \cite{Du2017Learning}.
\item \textbf{Transformer (Scialom)}: This is the state-of-the-art result for the AG-QG task, which adopts a transformer network with some extension. \cite{Scialom2019Self}.
\end{itemize}
We do not take \cite{Wang2019A} into comparison because their evaluation is done on a different test set and is not accessible.

The experimental results are shown in Table \ref{evaluation of different models}. The full version of our model which uses both the question type module and copy loss mechanism obtains the best results on all of metrics, achieving a new state-of-the-art result of BLEU-4 13.90 for the challenging AG-QG task. It outperforms the baseline model with 0.73 points and beats the previous best result by 0.67 points.

% By incorporating the copy mechanism, our baseline model achieves a high BLEU-4 score of 13.17, which outperforms the traditional seq2seq attention model by 0.68 points and so provides us a strong baseline. It demonstrates that the copy mode plays an important role for question generation. 

\subsection{Ablation Study}
We conduct extensive experiments with different model modules, where $k$ is set to 1 in decoding. The results are reported in Table \ref{ablation study}.
\begin{itemize}
 %implemented by us. To be specific, we do not use predict type module, and the hidden state at the first decoding step $h_0=LSTM(u_l,[SOS])$, where $[SOS]$ is the start of sequence symbol, and at every decoding step, $g_t=softmax(\mathbf{W^o}tanh(\mathbf{W^t}[h_t;v_t]))$. Besides, we do not use copy mechanism and only use the generate distribution for decoding.
%\item \textbf{Seq2SeqAtt+Type}: This is the model that applies question type predict module. That is, $h_0=LSTM(u_l,qt)$ and $g_t=softmax(\mathbf{W^o}tanh(\mathbf{W^t}[h_t;v_t;qt]))$. Copy mechanism is not used.
\item \textbf{Baseline}: Our baseline model is a general sequence-to-sequence attention model enhanced with copy mechanism.
\item \textbf{Baseline+Type}: It adds the question type module to the baseline model.
\item \textbf{Baseline+CopyLoss}: Based on the baseline model, it calculates and minimizes the additional copy loss.
\item \textbf{Baseline+CopyLoss+Type}: This is the full version of our proposed model. That is, the question type module is applied to the baseline model and the extra copy loss is also calculated. 
\item \textbf{Upper Bound}: Since our full model incorporates the question type prediction part, the accuracy of question type prediction will undoubtedly affect the final quality of generation. If the right question type is given for every test sample, we get the upper bound of our model.
\end{itemize}

% \subsubsection{Upper Bound Analysis} The upper bound achieves a Bleu\_4 of 15.27, which demonstrates the huge potential of our model, if the accuracy of question type prediction can be improved. 

\paragraph {Effect of Question Type Module.}
Comparing with the baseline model, the question type module brings a slight performance gain. The upper bound shows if the right type is given for each test sample, the model will yield a much better performance with a BLEU-4 score of 15.27, which demonstrates the huge potential of our model. It proves that our model has successfully learned the patterns of different types of questions.

However, our question type predict module cannot achieve a 100\% accuracy, and once a wrong question type is offered to the decoder, it will have negative influence on the generating quality. Actually, our question type predict part achieves an overall 69\% accuracy, and the prediction results of different question types are shown in Table \ref{question type predict}. It shows that without the answer as input, to predict the types of questions that should be asked for a given sentence is non-trivia.   

\begin{table}[t]
\small
\begin{tabular}{lp{0.7cm}p{0.9cm}p{1.1cm}p{0.9cm}p{1cm}}
\hline
\textbf{Type} & \textbf{Total} & \textbf{Predict} & \textbf{Precision} & \textbf{Recall} & \textbf{Fscore} \\
\hline
\hline
what & 6707 & 8542 & 0.63 & 0.80 & 0.71 \\
who & 1421 & 1298 & 0.35 & 0.32 & 0.34 \\
how & 1476 & 1258 & 0.59 & 0.50 & 0.54 \\
where & 455 & 350 & 0.19 & 0.15 & 0.17 \\
when & 647 & 346 & 0.27 & 0.15 & 0.19 \\
yes/no & 868 & 65 & 0.09 & 0.01 & 0.01 \\
others & 303 & 18 & $<$ 0.01 & $<$ 0.01 & $<$ 0.01 \\
\hline
\end{tabular}
\caption{Prediction results of different question types.}
\label{question type predict}
\end{table}

% The upper bound shows if right type is given for each test sample, the model achieves a much better performance, which proves that our model has successfully learned the patterns of different types of questions. However, our question type predict module cannot achieve 100\% accuracy, actually only 69\% accuracy, and once wrong question type is offered to decoder, it leaves negative influence on the generating quality. The result shows with a predict accuracy of 69\%, our model can get a stable 0.3 increment on Bleu\_4. 

\begin{table}[t]
\small
\begin{tabular}{lp{2.3cm}<{\centering}p{2.3cm}<{\centering}}
\hline
\textbf{} & \textbf{Key Words} & \textbf{Percentage}  \\
\hline
\hline
Baseline+Type & 1.26 &  40.78\% \\
Our model & \textbf{1.40} & \textbf{45.31\%} \\
\hline
Golden & 3.09 & 100\% \\
\hline
\end{tabular}
\caption{Performance of copying key words between two models with and without the enhanced copy mechanism.}
\label{copy key words}
\end{table}

\paragraph {Effect of Enhanced Copy Mechanism}
Our designed copy loss aims to enhance the copy mechanism. Since it tries to make the model ensure that every key word is copied, it directly leads to a higher BLEU-1. To our delight, the experiment shows the copy loss mechanism also contributes to a stable 0.48 increment of BLEU-4.

To make an in-depth analysis on the new copy mechanism, we also conduct experiments with and without the copy loss, counting the average number of keywords in questions generated by different models. The results are shown in Table \ref{copy key words}. Our copy loss brings an absolute 4.53\% increment on copying words from source sentences, which helps the model generate higher quality questions.    

\subsection{Human Evaluation}
We also conduct human evaluation to judge the quality of questions generated by our model and the baseline model, respectively. We take three metrics into consideration: 1) Fluency: it measures whether the question is grammatical; 2) Relevance: it measures whether the question is asked for and highly related to the source sentence; and 3) Answer-ability: it measures whether the generated question can be answered by the information of the source sentence. We randomly selected 100 sentence-question pairs generated by different models, and asked three annotators to score the questions on a $1\!-\!5$ scale (5 for the best). We also exploit the Spearman correlation coefficient to measure the inter-annotator agreement. The results are shown in Table \ref{human evaluation}. It shows that the consistency among three annotators is satisfying, and our generated questions are better from different perspectives.

Besides, a further two-tailed t-test proves that our generated questions are better than that of the baseline model significantly, with p $<$ 0.001 for every metric.

\begin{table}[t]
\small
\begin{tabular}{lp{1.55cm}<{\centering}p{1.55cm}<{\centering}p{1.55cm}<{\centering}}
\hline
\textbf{} & \textbf{Fluency} & \textbf{Relevance} & \textbf{Answer-ability} \\
\hline
Baseline & 3.78 & 3.73 & 3.54 \\
Our model & \textbf{4.23} & \textbf{4.23} & \textbf{4.20} \\
\hline
Golden & 4.69 & 4.44 & 4.48 \\
\hline
\hline
Spearman & 0.59 & 0.67 & 0.65 \\
\hline
\end{tabular}
\caption{Human evaluation results of different models.}
\label{human evaluation}
\end{table}

\begin{table}[t!]
\small
\begin{tabular}{|p{7.4cm}|}
\hline
\multicolumn{1}{|c|}{Sample 1} \\
\hline
\textbf{Source sentence}: the ex-president of def jam l.a. reid has described beyoncé as the greatest entertainer alive . \\
\hline
\textbf{Ground-truth}: \textbf{who} has said that \emph{beyoncé} is the best entertainer alive ? \\
\textbf{Baseline}: what is the greatest entertainer alive ? \\
\textbf{Our model(K=1)}: \textbf{who} described \emph{beyoncé} as the greatest entertainer alive ? \\
\hline
\hline
\multicolumn{1}{|c|}{Sample 2} \\
\hline
\textbf{Source sentence}: tibetan sources say deshin shekpa also persuaded the yongle emperor not to impose his military might on tibet as the mongols had previously done . \\
\hline
\textbf{Ground-truth}: \textbf{who} convinced the \emph{yongle emperor} not to send \emph{military} forces into \emph{tibet} ? \\
\textbf{Baseline}: what did tibetan sources say deshin ? \\
\textbf{Our model(K=1)}: \textbf{who} persuaded the \emph{yongle emperor} not to impose his \emph{military} might on \emph{tibet} ? \\
\hline
\end{tabular}
\caption{Examples of generated questions by different models.}
\label{case study}
\end{table}

%\begin{figure*}[t!]
%	\centering
%	\includegraphics[width=1\linewidth]{attention.jpg} 
%	\caption{Attention for different types of generated questions.}
%    \label{attention map}
%\end{figure*}

\subsection{Case Study}
In order to show the effectiveness of our model, we offer two real samples in the test set, as shown in Table \ref{case study}. In both samples, the baseline model generates a wrong type of question while our model predicts the right type of question. At the same time, our model successfully copies more key words from the source sentence, which are shown in italics, while baseline model fails. In both samples, our generated questions are more fluent and coherent.   

\begin{table}[t!]
\small
\begin{tabular}{|p{7.4cm}|}
\hline
\multicolumn{1}{|c|}{Sample 1} \\
\hline
\textbf{Source sentence}: by 1790 , new york had surpassed philadelphia as the largest city in the united states . \\
\hline
\textbf{Groud-truth}: \\
by \textbf{which} year , did new york city become the largest city in the united states ? \\
\textbf{what} was the second largest city in the united states in 1790 ? \\
\hline
\textbf{Our Model (K=2)}: \\
1st: in \textbf{what} year did new york city surpassed philadelphia ? \\
2nd: \textbf{which} city had surpassed philadelphia in the us ? \\
\hline
\hline
\multicolumn{1}{|c|}{Sample 2} \\
\hline
\textbf{Source sentence}: buddhist architecture , in particular , showed great regional diversity . \\
\hline
\textbf{Groud-truth}: \\
\textbf{which} cultures architecture showed a lot of diversity ? \\
\textbf{what} type of architectural is especially known for its regional differences ? \\
\hline
\textbf{Our Model (K=2)}: \\
1st: \textbf{which} buddhist architecture has showed great regional diversity? \\
2nd: \textbf{what} is buddhist architecture ? \\
\hline
\end{tabular}
\caption{Different types of questions generated from the same input sentence.}
\label{multi-types}
\end{table}

\begin{figure}[h]\centering
\includegraphics[width=.45\textwidth]{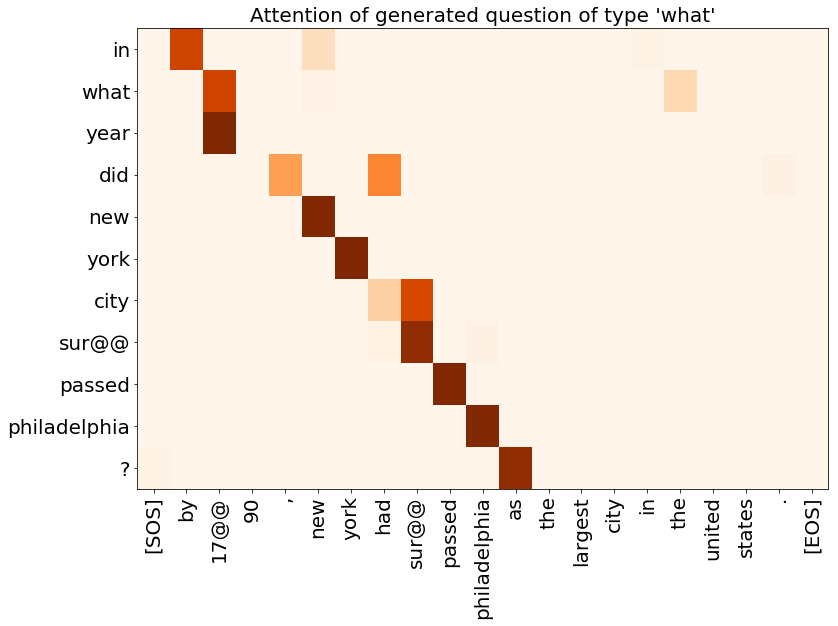}
\includegraphics[width=.45\textwidth]{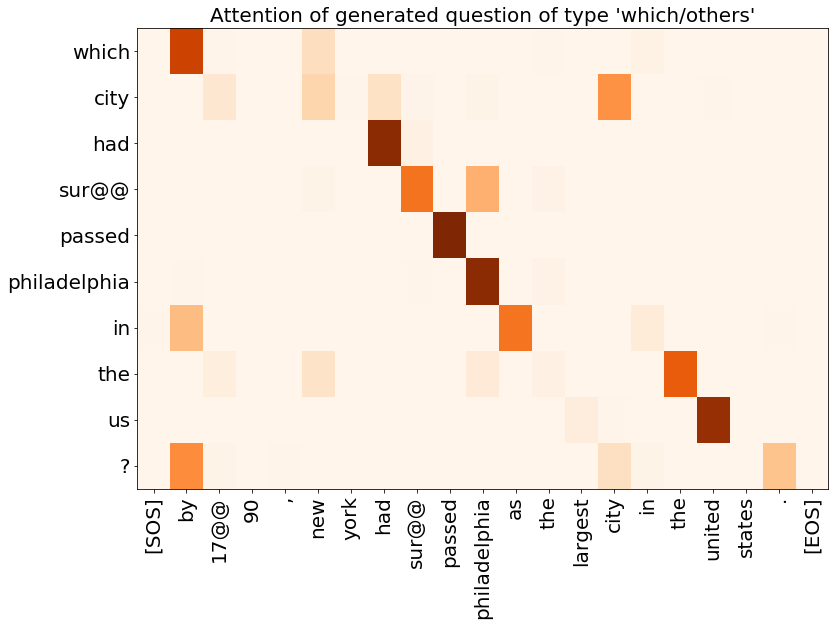}
\caption{Heat map of attention for different types of generated questions.}
\label{attention map}
\end{figure}

%\begin{figure*}[h]
%    \begin{tabular}{ccc}
%    \begin{minipage}[t]{0.33\textwidth}
%    \includegraphics[height=5.5cm,width=5.5cm]{which.png}
%    \end{minipage}
%    %%
%    \begin{minipage}[t]{0.33\textwidth}
%    \includegraphics[height=5.5cm,width=5.5cm]{what.png}
%    \end{minipage}
%    %%
%    \begin{minipage}[t]{0.33\textwidth}
%    \includegraphics[height=5.5cm,width=5.5cm]{how.png}
%    \end{minipage}
%    \end{tabular}
%    \caption{Attention for different types of generated questions.}
%    \label{attention map}
%\end{figure*}

\subsection{Asking Different Types of Questions}
For a given sentence, our question type driven framework offers the model the ability to generate different types of questions. In this case, the parameter $K$ is set to more than 1, and the question type predictor will give $K$ question types with the highest possibility. Then the model automatically decodes $K$ times to generate the best $K$ types of questions. We list a sample in Table \ref{multi-types} with $K=2$ to show the generating diversity of our model, where two types of questions (what and which) are generated from the same input sentence. 

Besides, to identify the effect of our model, we visualize the decoder attention, as shown in Figure \ref{attention map}. The two attention maps show the attention points when our model is generating different types (which and what) of questions with respect to the same input sentence, where $x$-axis is the source sentence and $y$-axis is the generated question. Differences between these attention maps prove that our model can attend on different information when generating different types of questions. 

From the table, we prove that our model has the ability to generate multiple questions. However, the limitation is also obvious. First, if $K$ is too large, the generated questions of some low probable types are of low quality. 
% for example, in this case, the question of type 'who' is almost the same as the question of type 'what', only the question-word is replaced, and the question is illogical. Similar phenomenon also occurs between generated questions of type 'when' and 'where', that question of type 'when' is of high quality while question of type 'where' is illogical. 
Second, since the probability distribution of question types are automatically calculated, the types of generated questions cannot be known beforehand.
%, which we classify as type 'others'(which, why and so on). That our model fails to generate good questions of types other than 'what' to some extend attributes to the fact that more then 60\% ground-truth in SQuAD are of type 'what'. Fortunately, our question type predict part successfully predicts the right type, 'what', which is also the most proper one in this case.

% \subsection{Question Type Predict Accuracy}
% In this part we briefly evaluate the performance of our question type predict part, result is showed in table\ref{question type predict}. Notice that since many source sentences are offered different types of questions and we can only predict one type for one source sentence, the sum of the predict number is less than the sum of total number. 

% Our question type predict part achieves an overall 69\% accuracy, which is not satisfying. And the accuracy of each type is severely imbalance. The most important reason is the imbalance of data distribution. Besides, we suggest the poor accuracy of type 'yes/no' and 'others' is because that we mix more than one specific types into these two category, which makes its pattern confusing and hard to classify. But even though the accuracy of question type prediction is unsatisfying, it still helps our model to outperform the previous state-of-the-art.

\section{Conclusion}
In this paper, we propose two new strategies to deal with the answer-agnostic QG: question type module and copy loss mechanism. 
% modules: penalty loss mechanism and question type module, which actually consists of a question type embedding part and an optional question type predict part. 
These proposed modules improve the performance over the baseline model, achieving the state-of-the-art. Moreover, our model has the ability and flexibility to generate multiple questions for one source sentence. Hopefully, the idea of question type module and copy loss mechanism can also be used to do answer-aware QG task or other similar text generation tasks.
%However, our work still has limitations. First, the multiple questions generated for one source sentence are of low diversity, that means, although they are of different types, they actually look similar with each other. It is because our question type module is still simple, and we are seeking more complex alternative to address this problem. Secondly, the unsatisfying accuracy of the question type predict part prevents our model to achieve a better performance. Predicting question type without answer information is very hard but valuable since higher accuracy can definitely lead to performance that is closer to the upper bound. Hopefully, the idea of question type module and penalty loss mechanism can also be used to do answer-aware QG task.

\section*{Acknowledgments}
This work is supported by the National Natural  Science Foundation of China (61773026) and the Key Project of Natural Science Foundation of China (61936012).

\bibliography{acl2020}
\bibliographystyle{acl_natbib}

\end{document}